**Title**:
Deep reinforcement learning to detect brain lesions on MRI: a proof-of-concept application of reinforcement learning to medical images

**Authors:**
Stember JN[1*], Shalu H[2]

*Corresponding author

1. Memorial Sloan Kettering Cancer Center
Department of Radiology
1275 York Avenue
NY, NY 10065

2. Indian Institute of Technology Madras
Department of Aerospace Engineering
IIT P.O., Chennai 600 036
INDIA

**Abstract**

*Purpose*

Artificial intelligence (AI) in radiology is hindered chiefly by: 1) Requiring large annotated data sets. 2) Non-generalizability that limits deployment to new scanners / institutions. And 3) Inadequate explainability and interpretability, these being critical to foster the trust needed for wider clinical adoption.

We believe that Reinforcement Learning can address all three shortcomings, with robust and intuitive algorithms trainable on small datasets. We in fact feel that reinforcement learning will help to usher radiology AI beyond its infancy and into true clinical applicability, while broadening its research horizons. To the best of our knowledge, reinforcement learning has not been directly applied to computer vision tasks for radiological images. In this proof-of-principle work, we train a deep reinforcement learning network to predict brain tumor location.

*Material and Methods*

Using the BraTS brain tumor imaging database, we trained a deep Q network (DQN) on 70 post-contrast T1-weighted 2D image slices. We did so in concert with image exploration, with rewards and punishments designed to localize lesions.

To compare with supervised deep learning, we trained a keypoint detection convolutional neural network on the same 70 images. We applied both approaches to a separate 30 image testing set.

*Results*

Reinforcement learning predictions consistently improved during training, whereas those of supervised deep learning quickly diverged. Reinforcement learning predicted testing set lesion locations with 85% accuracy, compared to roughly 7% accuracy for the supervised deep network.

*Conclusion*

Reinforcement learning predicted lesions with high accuracy, which is unprecedented for such a small training set. We have thus illustrated some its tremendous potential for radiology. Most of the current radiology AI literature focuses on incremental improvements in supervised deep learning. We believe that reinforcement learning can propel radiology AI well past this inherently limited approach, with more clinician-driven research and finally toward true clinical applicability.

## Introduction

Over the past few years, artificial intelligence (AI) and deep learning have gained traction and ubiquity in radiology research. Indeed, the volume of research in this field has grown exponentially (1–3).

However, significant hurdles remain, and the key to overcoming them may lie in areas of AI hitherto unexplored in radiology. A 20,000-foot view of deep learning is in order: most radiology AI up until now has been in the realm of supervised deep learning (SDL). In SDL, large numbers of explicitly labeled data are used to train convolutional neural networks (CNNs). These trained networks can then make predictions on new unlabeled data, often classification, semantic segmentation or localization tasks.

SDL is complemented by unsupervised deep learning, which seeks to cluster data without pre-labeling by an imaging expert. The goal in unsupervised deep learning is to uncover commonalities and differences in data, without overt data labeling or annotation. Unsupervised deep learning is more of an exploratory procedure; typically, the researcher does not initially know what s/he is looking for specifically, and no goal can be provided to the algorithm. Unsupervised learning is a small but growing portion of the landscape (4,5). However, SDL continues to constitute the vast majority of current radiology deep learning research.

SDL suffers from the following three major limitations:
1. It requires tremendous volumes of expertly hand annotated data, which is time-consuming and tedious.

2. It is unstable in the sense of being exquisitely sensitive to even subtle differences in image quality. Goodfellow et al. showed that adding just a small amount of imperceptible image noise can throw off the predictions of SDL (6). As such, a network trained on a given patient population with certain CT or MRI scanner settings at a particular institution in general fails to perform acceptably well when deployed on a new scanner and/or different institution with different patient population.
3. SDL suffers from a notorious “black box” phenomenon (7,8). Gaining the trust of people, particularly in a field as critical as health care, requires that the algorithms helping to make decisions do so in a transparent manner. Understanding at least part of the rationale for why AI algorithms make certain determinations fulfills this basic need of both patients and clinicians that is required before widescale adoption is possible (9,10).

In our search to overcome these hurdles, we note that a third major field of AI research has already produced astounding results in applications as varied as board games, automated video game playing and robotics. This AI field is called reinforcement learning (RL), and it is the focus of the present work. RL lies somewhere in between supervised and unsupervised deep learning, yet is in important ways distinct from both (5). Notably, RL was a key contributor to AlphaGo’s victory over the European champion in the game of Go, a major breakthrough given the game’s inherent complexity (11). Additional advances that would have been unfeasible with supervised or unsupervised approaches include world champion level autonomous video game playing (12–14) and in the field of robotics (15,16).

RL can address the three aforementioned drawbacks of SDL:
1. The way RL learns about the environment / image includes much of what is not the desired label or structure of interest. As such, many fewer labeled images are needed for RL to learn enough to apply to new images.
2. Because of this learning of the “non-answer,” as opposed to SDL’s reliance on labeled correct answers, RL is robust to inevitable noise and variations in image acquisition techniques and patient populations. This has to do with how RL works; goals are not provided explicitly as in SDL, but rather implicitly through a system of rewards. This process turns out to provide a more robust and generalized kind of learning during training.
3. The reward system also provides vital intuition that is lacking in SDL. Reward structures provide the rationale for why an algorithm makes certain predictions. It also opens opportunities for imaging medical specialists such as radiologists and pathologists to exploit their domain knowledge to help craft algorithms.

Despite the promise, RL has not as of yet, to the best of our knowledge, been applied directly to computer vision in radiological / medical images. Coming closest, in the field of computer vision more generally, Wang and Sarcar were able to accurately segment non-radiological images using RL with simulated pen drawings (17).

We seek in this work to demonstrate proof-of-principle application of RL to radiology. Our application is to detect brain tumors from 2D MRI images. This will lead to more sophisticated implementations for image classification, object detection and segmentation. We believe that RL will ultimately far exceed the accomplishments of SDL. We anticipate further that as RL in radiology advances in subsequent work, it will finally begin to demonstrate, in a paradigm-shifting manner, the truly enormous potential of AI to fundamentally improve the efficiency and success of clinical image interpretation.

Before describing the application of RL to our detection problem, we very briefly introduce the approach, starting with its history. RL gradually came into being through two disparate threads of research: behavioral psychology and engineering control theory. The notion of rewards and punishments engendering learning has a long history in psychology, originally formulated by Thorndike in 1911 based on animal studies (18) as the "Law of Effect." Minsky (19) and Bellman (20–22) provided the foundational control theory formulations in the 1950s. These two threads coalesced during a revival of the field in the 1980s–1990s. It was in this era that most of the modern RL concepts were developed by Barto and Sutton (23–27). More recently, the introduction of deep convolutional neural networks has produced deep reinforcement learning (DRL). The associated deep Q-networks (DQNs) provide functional approximations that can ultimately allow for optimal actions to be taken. For example, an early success for DRL was champion-level performance on Atari games (12) because it allowed for the autonomous game-player to select the optimal action at every step of the game based purely in pixelwise input. We employ a DRL formulation in the present work.

## Methods

### *Concepts and terms*

In order to describe the method, we need to very briefly introduce some key concepts from RL/DRL. We will do so in the context of our particular system and approach. For the interested reader, much more detailed and didactic treatments of RL and DRL are available elsewhere, for example the textbook by Barto and Sutto (28).

RL focuses on the process of learning from an environment (28,29). This type of learning is guided by the experiences of an agent, which is able to take certain actions within the environment. Depending on the action and particular state, the agent receives certain rewards. Once one has specified the environment, states and possible actions, a careful selection of the rewards can guide learning to fulfill a desired task. The goal of the algorithm, or agent, is to achieve the maximum cumulative reward. This can be at the cost of short-term gains.

More formally defining our terms:
-Environment: This is the physical world in which the agent operates and with which it interacts. In our case, the environment consists mostly of the 2D slice of a post-contrast T1-weighted brain MRI containing a glioblastoma multiforme (GBM) lesion. In order to make the problem of finding this lesion more tractable, we add to our environment the set of (x,y) positions looked at by a radiologist at an earlier time during simulated image interpretation, as recorded by eye tracking hardware and software (30–34). This set of (x, y) coordinates within the image is referred to as the gaze plot. Hence the full environment here is the 2D image with overlaid gaze plot for that image.
-Agent: entity that takes actions in an environment and receives rewards. In our case the agent is a moving point sitting on a pixel that we ultimately hope will land within the tumor, thereby predicting its position.
-State: the state conveys the agent's current situation. In our case the state is where in the image our point resides, i.e. on which pixel it is currently sitting. In order to decrease the state space, or possible ways the point can move as it seeks to find the lesion, we restrict our agent to move only along the gaze plot, i.e. among pixels that were looked at by the radiologist. The assumption is that during the prior simulated image interpretation, the radiologist (JNS, with two years of experience in neuroradiology) looked at the lesion at least once. Hence, by moving along gaze points, which constitutes a one-dimensional space, the agent is certain to intersect the lesion. How the agent can move between states is illustrated in Figure 1.
-Action: a change in the state. In our case, action is our point moving between pixels in the image, more specifically between points on the gaze plot.
-Policy: the prescription for which action to take in a given state. The goal of RL/DRL training is to produce an optimal policy. In our case, the optimal policy is for the agent to move toward the tumor as quickly as possible and stay there to mark / predict that lesion.
-Value: the cumulative future reward that the agent receives by taking a given action in a particular state. In our system, moving toward the lesion has high value, whereas moving away from it has low value.

It should be noted that our environment satisfies the property of being a Markov Decision Process. Markov Decision Processes are environments in which essentially all RL problems can be formulated. Their chief attribute is lack of prior knowledge of environment dynamics.

*Rewards*

We build our reward system so as to align the agent's goal of maximizing cumulative reward with our objective to detect brain lesions. We wish to incentivize reaching the pixels in the tumor and then staying within the tumor, and de-incentivize staying still outside the tumor or moving away from the tumor. To this end, the reward $R$ is defined by:

$$R = \begin{cases} +2, \text{if agent is within lesion and staying still} \\ -4, \text{if agent is outside lesion and staying still} \\ +0.5, \text{if agent is within lesion and moving backward} \\ -1.5, \text{if agent is outside lesion and moving backward} \\ +0.5, \text{if agent is within lesion and moving forward} \\ -0.5, \text{if agent is outside lesion and moving forward} \end{cases} \quad (1)$$

As stated above, we restrict our agent's possible positions to being along the 1D gaze plot. We do so in order to decrease the state and action space and simplify learning calculations. We define the anterograde direction as being toward the final point in the gaze plot, which is the last point at which the radiologist had looked for that 2D image. Retrograde is toward the initial point, looked at first by the radiologist.

*Actions*

From the current state / gaze point, we define 3 possible actions that the agent can take:
1) Moving anterograde (if not at the last gaze point, in which case it stays still).
2) Not moving
3) Moving retrograde (if not currently on the first gaze point, in which case it would not move).
In other words, the action vector $A \equiv \{\rightarrow, \circlearrowleft, \leftarrow\}$, where $\rightarrow$ denotes moving anterograde along the gaze plot, $\circlearrowleft$ is staying still in the same pixel and $\leftarrow$ is moving retrograde. The agent is only allowed to move by one gaze point at a time.

*States and policy*

Our state space is defined by where along the gaze plot we are, i.e. on which point our agent is located. We denote the gaze plot for the $j$th image $I_j$ consisting of $N_{\text{gaze}}^{(j)}$ points, by $\left\{g_i^{(j)}\right\}_{i=1}^{N_{\text{gaze}}^{(j)}}$. Our agent begins all training episodes at the first point $g_1^{(j)}$. This state is displayed in Figure 2. The state consists of the 2D brain MRI slice overlaid with gaze plot points in red, then which gaze point the agent is currently located on as a blue square of size 11 x 11 square pixels centered on the point. In order to allow the algorithm to "see through" the gaze points and square representing agent position, both were set to partial transparency. However, the figures here show them with no transparency for the sake of display.

In order to learn the optimal policy $\pi$, we define a state-action value or quality function $Q^\pi(s, a)$, as the expected / average total reward when taking action $a$ in state $s$ and and then all actions according to policy $\pi$ thereafter:

$$Q^\pi(s, a) = \mathbb{E}[R | s, a, \pi], \tag{2}$$

Where $\mathbb{E}$ is the expectation value. Now if we could learn $Q$ for all possible state-action pairs and find a policy that maximizes $Q$, we could follow that policy by picking best actions to arrive at the lesion quickly and reliably. In order to do so, $Q$-learning can be employed, wherein $Q$ values for each possible station-action combination are calculated in an iterative fashion, eventually converging to true values. The problem with $Q$ learning is that it is unfeasible for all except small systems with few possible states and actions. Instead, for most applications, including ours, it is preferable to learn a function approximation of $Q$, as a function of state and action. That allows us to calculate $Q$ values for states and actions not yet seen in iterative exploration, but interpolated from nearby values that were sampled.

*Deep Q Network*

As is known from SDL, CNNs provide excellent function approximation. We can similarly employ them to learn $Q$-functions, and call them deep $Q$-networks (DQNs). The architecture of our DQN is shown in Figure 2. Taking the state as input, we used 3 x 3 kernels with stride 2 and padding so as to maintain the size of the resulting filters. We produced 32 filters at each convolution operation. The network consisted of four such convolutional filters in sequence, using exponential linear unit (elu) activation. The last convolution layer was followed by a fully connected 512-node layer, fully connected to a 3-node output layer, representing 3 $Q$ values, one corresponding to each possible action. Our loss is the difference between the $Q$ values resulting from a forward pass of the network, which we shall denote as $Q_{\mathrm{CNN}}$ and the "target" $Q$ value computed by the Bellman equation, which updates by sampling from the environment and experiencing rewards.

*Bellman Equation*

The Bellman update equation is given by:

$$Q(s_t, a_t) \longleftarrow Q(s_t, a_t) + \alpha[r_t + \lambda max_a Q(s_{t+1}, a_t) - Q(s_t, a_t)], \qquad (3)$$

where $\propto$ is the learning rate and $\lambda$ is the discount factor, which reflects the present value of future rewards, the latter factored more as $\lambda$ is increased toward one. $Q(s_t, a_t)$ is the current $Q$ value being updated and $max_a Q(s_{t+1}, a_t)$ is the estimated reward from our next action given on-policy behavior, i.e. taking actions that maximize cumulative future reward.

As discussed more below, using a form of equation (3), we update $Q$ values, and record these along with the state, action and reward values, calling the result tuple a transition, $\mathrm{T}_t = (s_{t,} a_t, r_t, s_{t+1})$.

*Workflow*

Our workflow proceeds as follows:
Starting with the first gaze point, $g_1^{(j)}$, we create a square matrix around that pixel which is 11 x 11 $\text{pixels}^2$. This size is chosen to stand out as a state more than just the single point, and thus we expect the DQN to "see it" better. Nevertheless, the square represents the single center point and we will interconvert between the two.

We next pursue an interleaved process of sampling and learning from our environment using the reward scheme from Equation 1, sampling the state-action space via the Bellman Equation and training our DQN. Again, doing the first part ensures that we will obtain a $Q$ estimation that approaches $Q*$, the optimal policy. The DQN training assures that we have a function approximation for $Q$ that also approaches $Q*$. Hence, in the end we have a function taking states as input and calculating $Q$ values of the three possible actions. We can then select the optimal action to take as simply the action giving the largest $Q$ value, i.e. the argmax over actions. Then we can optimally move the agent point around the image so as to ensure that at the end of a testing set episode the point is inside the lesion. Thus, we will have localized the tumor.

We use a training set of 70 images + gaze plots. We select each image at random and compute the square centered at the first gaze point. We overlay the corresponding pixels in blue onto the 2D grayscale image slice already overlaid with gaze points in red. We plot red and blue representing gaze points and agent locations, respectively, with partial transparency. This image serves as the initial state, $s_1$ (Figure 2, no transparency for purposes of display). Next we select an action $a_1$ to perform on this state based on the $\epsilon$-greedy algorithm: if a random number between zero and one is larger than $\epsilon$, then an on-policy action is selected. This action is computed by acting on $s_1$ with a forward pass of the DQN, denoted in functional form as $\mathcal{F}_{\mathrm{CNN}}$. The forward pass $\mathcal{F}_{\mathrm{CNN}}$ outputs three nodes, one corresponding to the $Q$ value of each possible action (Figure 2). Then picking the argmax action corresponding to the largest $Q$ value, we choose the optimal / on-policy action for this step. If, on the other hand, the random number is less than $\epsilon$, then the action is chosen at random from among the three possible actions.

Of note, initially the policy is purely random because we do not know from the outset what the policy actually is. This is manifested as DQN's weights being initialized according to a Glorot random distribution. Our algorithm learns the optimal policy through repeated iterations of exploring its environment.

Upon taking the selected action $a_1$, our agent arrives at new state $s_2$, receiving an award $r_1$ as per Equation 1. In this case of the initial state, we record the transition $\mathrm{T}_1 = \left(s_{1,} a_1, r_1, s_2\right)$. In general, we record $\mathrm{T}_t = \left(s_{t,} a_t, r_t, s_{t+1}\right)$. As we continue running the process, we keep adding rows $\mathrm{T}_t$ to an initially growing transition matrix $\mathbb{T}$, up to a maximum number of 12,000 transitions stored as 12,000 rows in $\mathbb{T}$. This number of rows / transitions is specified by the memory size $N_{\mathrm{memory}}$. Once we have added enough transitions $\mathrm{T}_t$ to bring $\mathbb{T}$ to its maximum size, $N_{\mathrm{memory}} \times 4 = 12{,}000 \times 4$, we begin discarding the earliest transitions as new ones are added, keeping the number of rows fixed at $N_{\mathrm{memory}}$. For example, once we add the 12,001st transition $\mathrm{T}_{12{,}001}$, we have to remove the first row $\mathrm{T}_1$. Then when adding $\mathrm{T}_{12{,}002}$ to $\mathbb{T}$, we make room by removing $\mathrm{T}_2$.

We note that $N_{\mathrm{memory}}$ is an adjustable hyperparameter. The tradeoff in selecting the best memory size value: larger values of $N_{\mathrm{memory}}$ yield better DQNs because more transition samples are used for its training, learning more about the environment. However, larger $N_{\mathrm{memory}}$ slows the calculation and at a certain point will overwhelm the CPU's random access memory (RAM).

Going back to the start of our procedure, having just calculated $\mathrm{T}_1$, we next compute a target $Q$ value, $Q_{\mathrm{target}}^{(1)}$ by the Bellman Equation as $r_1 + \lambda max_a Q(s_2, a)$, where $max_a Q(s_2, a)$ is the action of the maximum $Q$ output from the forward feed of the CNN at $s_2$, i.e. $\mathcal{F}_{\mathrm{CNN}}(s_2)$. In general, with the following modified Bellman equation, we calculate

$$Q_{\mathrm{target}}^{(t)} = r_t + \lambda max_a Q(s_{t+1}, a). \tag{4}$$

We also calculate $Q_{\mathrm{CNN}}^{(t)} = \mathcal{F}_{\mathrm{CNN}}(s_t)$, where $\mathcal{F}_{\mathrm{CNN}}$ again is the function / operator of a forward pass of the DQN. Then we have a vector / set of $N_{\mathrm{memory}}$ =12,000 target values $Q_{\mathrm{target}} = \left\{Q_{\mathrm{target}}^{(t)}\right\}_{t=1}^{N_{\mathrm{memory}}}$, and 12,000 DQN-predicted values, $Q_{\mathrm{CNN}} = \left\{Q_{\mathrm{CNN}}^{(t)}\right\}_{t=1}^{N_{\mathrm{memory}}}$. In each step backpropagating our DQN, we randomly select a batch size of $N_{\mathrm{batch}} = 64$ transitions and then backpropagate to minimize the loss $\mathcal{L}_{\mathrm{batch}}$ of that batch,

$$\mathcal{L}_{\mathrm{batch}} = \frac{1}{N_{\mathrm{batch}}} \sum_{i=0}^{N_{\mathrm{batch}}} \left| Q_{\mathrm{target}}^{(i)} - Q_{\mathrm{CNN}}^{(i)} \right|. \tag{5}$$

Having backpropagated and re-adjusted DQN weights, we then take our next action and update $\mathbb{T}$, then re-compute $Q_{\mathrm{target}}$ and $Q_{\mathrm{CNN}}$. We randomly select another batch, recompute $\mathcal{L}_{\mathrm{batch}}$ and run another backpropagation, again updating the DQN weights. We continue in this manner for all the steps in each successive episode. By this process, $Q_{\mathrm{target}}$ approaches the optimal value $Q^*$ as we continue to sample our environment, and $Q_{\mathrm{CNN}}$ approaches $Q_{\mathrm{target}}$ as our DQN learns to minimize the loss. Hence, $Q_{\mathrm{CNN}}$ approaches $Q^*$, and we ultimately reach the optimal policy.

The images we used for training and testing came from the BraTS high grade glioma database (35). From that database's T1-weighted post-contrast 3D image volumes, we randomly selected 100 2D image slices. We employed a 70/30 training-testing split, using the first 70 of these images for training. We trained for 300 episodes, where an episode was defined as running $N_{\mathrm{gaze}}^{(j)}$ steps of simulation on image $I_j$. We note that the analogue of number of episodes in DRL is number of epochs for SDL.

Our DRL agent had input parameters as follows:
$\gamma = 0.99$, to reflect the fact that we wanted our agent to count future rewards significantly.
$\epsilon = 0.5$, with a rate of decay of $1 \times 10^{-4}$, so that $\epsilon$ would be slowly decreased by this amount each episode until reaching what we defined as a minimal value of $\epsilon_{\mathrm{min}} = 1 \times 10^{-4}$. This way, a lot of exploration could take place early in the training. This would give way to a steadily decreasing amount of random-move exploration as the algorithm learned the details of the images and $Q$ converged on the true optimal $Q*$. In other words, as the optimal policy was implicitly learned, the agent acted more and more according to that policy, while always leaving a little room to explore possible new solutions. We used a learning rate for our DQN of $1 \times 10^{-4}$, a standard-range value in SDL CNNs.

**Results**

Figure 3 displays a value for the agent's score during training. The score is the mean reward the agent receives during a given episode $i$, which for image $I_j$ is defined to last for a duration of $N_{\text{gaze}}^{(j)}$ steps:

$$\text{score}_i = \frac{1}{N_{\text{gaze}}^{(j)}} \times \sum_{k=1}^{N_{\text{gaze}}^{(j)}} R_k. \tag{6}$$

Figure 3 shows overall consistent improvement during training, although the inherent noisiness of the training process is apparent, with prominent jumps between episodes.

We trained for a total of 300 episodes, after which convergence was manifest, as can be seen in Figures 4 and 5. The training time was roughly 7 minutes and 31 seconds. Figures 4 and 5 show the training and testing set accuracies, respectively during the training process. We define a true positive training set state or testing set prediction as the agent's location being within the hand annotated lesion mask. We computed both training and testing values every 10 episodes simply as the proportion of number of true positives (TP) over those episodes:

$$\text{accuracy} = \frac{TP}{10}. \tag{7}$$

Figure 4 also shows ongoing improvement in training despite some noisiness. Importantly, Figure 5 tells us that we are not overfitting the training data, since we see a concurrent improving accuracy in localizing testing set lesions, on which no training was performed. The mean of the last 10 predicted testing set accuracies is 85%.

In order to compare DQN to SDL, we trained a keypoint detection CNN on the training set for 300 epochs using a network architecture that was identical to that of the DQN except for the last layer. For the keypoint detection network, the latter consisted of two nodes, followed by sigmoid activation, to represent the predicted x and y coordinates. In order to keep the comparison as close as possible, we trained the SDL CNN on the same training set of 70 images used for the DQN. We computed a loss of the testing set (sometimes alternatively in the context of an SDL CNN referred to as a validation set), shown along with the training set loss in Figure 6. The loss here is defined as negative mean absolute error from the center of the hand annotated mask's bounding box. Importantly, Figure 6 spotlights how the SDL CNN is already overfitting the training set by the 10$^{\text{th}}$ epoch, manifested by divergence of training and testing set losses.

We made DRL predictions on the testing set by applying our DQN to each of the 30 images and computing the network's accuracy. We act now according to the policy our DQN has implicitly learned, this approaching the optimal policy at later stages of training. At each step of training, for each testing set image $I_j$ with $j \in [71,100]$, we apply $\mathcal{F}_{\text{CNN}}$ and on-policy action selection for $N_{\text{gaze}}^{(j)}$ steps, starting with $s_1$. In general, at time step / iteration $t$, we get the three possible $Q$ values, corresponding to the three possible actions $\{a_{i,t}\}_{i=1}^{3}$, by applying a forward pass of the DQN:

$$\mathcal{F}_{\text{CNN}}(s_t) = \{Q_{a_{i,t}}\}_{i=1}^{3} = \begin{pmatrix} Q_{a_{1,t}} \\ Q_{a_{2,t}} \\ Q_{a_{3,t}} \end{pmatrix}. \tag{8}$$

Then we take the softmax $\sigma$ of the predictions, producing a probability distribution that sums to one:

$$\sigma(\mathcal{F}_{\text{CNN}}(s_t)) = \sigma\left(\{Q_{a_{i,t}}\}_{i=1}^{3}\right) = \begin{pmatrix} \frac{e^{Q_{a_{1,t}}}}{\sum_{k=1}^{3} e^{Q_{a_{k,t}}}} \\ \frac{e^{Q_{a_{2,t}}}}{\sum_{k=1}^{3} e^{Q_{a_{k,t}}}} \\ \frac{e^{Q_{a_{3,t}}}}{\sum_{k=1}^{3} e^{Q_{a_{k,t}}}} \end{pmatrix}. \tag{9}$$

Again, we are now acting exclusively on-policy, and compute the next action $a_t = \text{argmax}\left(\sigma\left(\{Q_{a_{i,t}}\}_{i=1}^{3}\right)\right)$. Then taking action $a_t$ brings us to state $s_{t+1}$. We continue in this manner until eventually reaching the last state $s_{N_{\text{gaze}}^{(j)}}$ after the $N_{\text{gaze}}^{(j)\ \text{th}}$ application of $\mathcal{F}_{\text{CNN}}$ and optimal action selection. If the agent's position in the final state $s_{N_{\text{gaze}}^{(j)}}$ lies within the lesion mask, the true positive count is increased by one. After computing the sum of true positives, $TP$, the accuracy is calculated as $TP/_{30}$.

For direct comparison between RL and SDL, we computed the mean of the last 10 predictions of both approaches as in Equation 7 (epochs or episodes 200, 210, …, 300 for SDL or RL, respectively). Doing so, the mean accuracy for RL was 0.85 (between 25 and 26 out of 30 lesions correctly predicted), while that for SDL was roughly 0.07 (2 out of 30 lesions). The difference, not surprisingly was statistically significant, with a $p$-value of $1.2 \times 10^{-22}$. The comparison of method predictions is shown in Figure 7.

**Discussion**

Although DRL is widely used in many cutting edge applications, such as robotics and game playing (36), to our knowledge it has not been applied directly to radiological computer vision. This work represents an initial proof-of-principle application of DRL/RL to MRI brain images. We have applied the approach to localize glioblastoma multiforme brain tumors from the BraTS public image database.

In so doing, we are able to glimpse the enormous potential of the DRL for radiology computer vision. We believe that this approach will in fact ultimately prove to be a seismic shift forward in artificial intelligence, one that will herald in the age of true clinical applicability, meaningfully changing the practice of radiology.

Training with DRL on only 70 training images, the algorithm was able to predict lesion location with 85% accuracy on a separate testing set. By comparison, this represents the general range of success for SDL when trained on hundreds or thousands of training set images, often with data augmentation. When trained on the same training set, the SDL network here quickly began to overfit the training data. It ultimately predicted testing set lesion location with accuracy of less than 10%. This last result is not surprising, given the widely known requirement of SDL for large amounts of annotated data to perform reasonably well.

We feel that we can confidently proclaim that DRL will change AI in radiology for two main reasons. It can:
1. Produce accurate results even when trained on very small data sets.
2. Provide and benefit from user intuition about the images being studied.

Regarding the first benefit, we note that 85% accuracy is a truly remarkable result for a training set of 70 images. Of note, we did not even employ data augmentation, a standard way to increase training set size used for SDL. Along with the obvious benefits of accelerating AI research by now allowing for small data sets to power effective AI, this ability provides two other key advantages.

Firstly, it allows for AI to accelerate and automate processing and analysis of the types of small datasets often encountered in academic settings. One may imagine for instance a particularly rare disease entity that only a handful of patients around the world have. The academic institution where we could imagine the foremost clinician treating perhaps 100 such patients would now be able to unleash the power of AI with even the correspondingly small collection of images.

Secondly, a decisive barrier to routine clinical application of current SDL-based algorithms is the drop-off in predictive power when a CNN trained on one data set is deployed for example at a new hospital, with the subtle variations secondary to different scanners with discrepant settings as well as the different patient populations. With its ability to train effectively on very small data sets, we expect that RL can be relatively easily and effectively retrained for new scanners / institutions.

The intuition for why RL is able to train effectively and in a generalizable fashion on such small data sets lies in the fact that it combs so extensively through the images. Whereas SDL learns what to do (e.g. locate or segment the lesion), RL learns what not to do (do so for non-lesion pixels). By learning all of the non-lesion parts of the image, the algorithm gains a more extensive sense of the image. It is thus less sensitive to the random noise that is introduced when encountering images from a new scanner.

As mentioned above, RL also provides intuition. This can alleviate the dreaded "black box" problem of SDL, in which limited-to-no information for why a network may be working or not, and what it is uncovering or what features it is geared for are hidden in the depth of the huge network, with its often millions of weights. By contrast, as illustrated in our example application, RL lets the user guide training by specifying appropriate rewards for certain kinds of actions in certain states. This offers understanding as to why a particular RL network is working or how it may be adjusted to work better. It also empowers radiologists and clinicians to bring their domain knowledge to bear on AI imaging training and applications.

Our system provides a nice illustration for this sort of intuition. Here, we want the agent not to "stand still" if it is not already in the lesion. Especially toward the onset near the initial gaze point, we want our agent to get moving and explore the state space so that it can find the tumor. Hence, we penalize the action of not moving while outside the tumor by a relatively high negative reward. Since we know that the goal is to localize tumors, we can tell our agent and hence DQN to usually stay still in the lesion once inside, by providing a generous reward to do so in training. Yet we still want it to explore a bit and find other points both within the lesion and "downhill" from it, so we give positive values for continuing to move within the lesion, though of lower magnitude than staying still in the tumor.

Future work will extend the approach to images without using eye tracking points, hence going from a one-dimensional state and action space to two dimensions and ultimately three-dimensions for full volumetric image stacks. Though this will increase the size and complexity of the state and action space, we will employ more sophisticated DRL techniques appropriate to these environments, such as the actor-critic method.

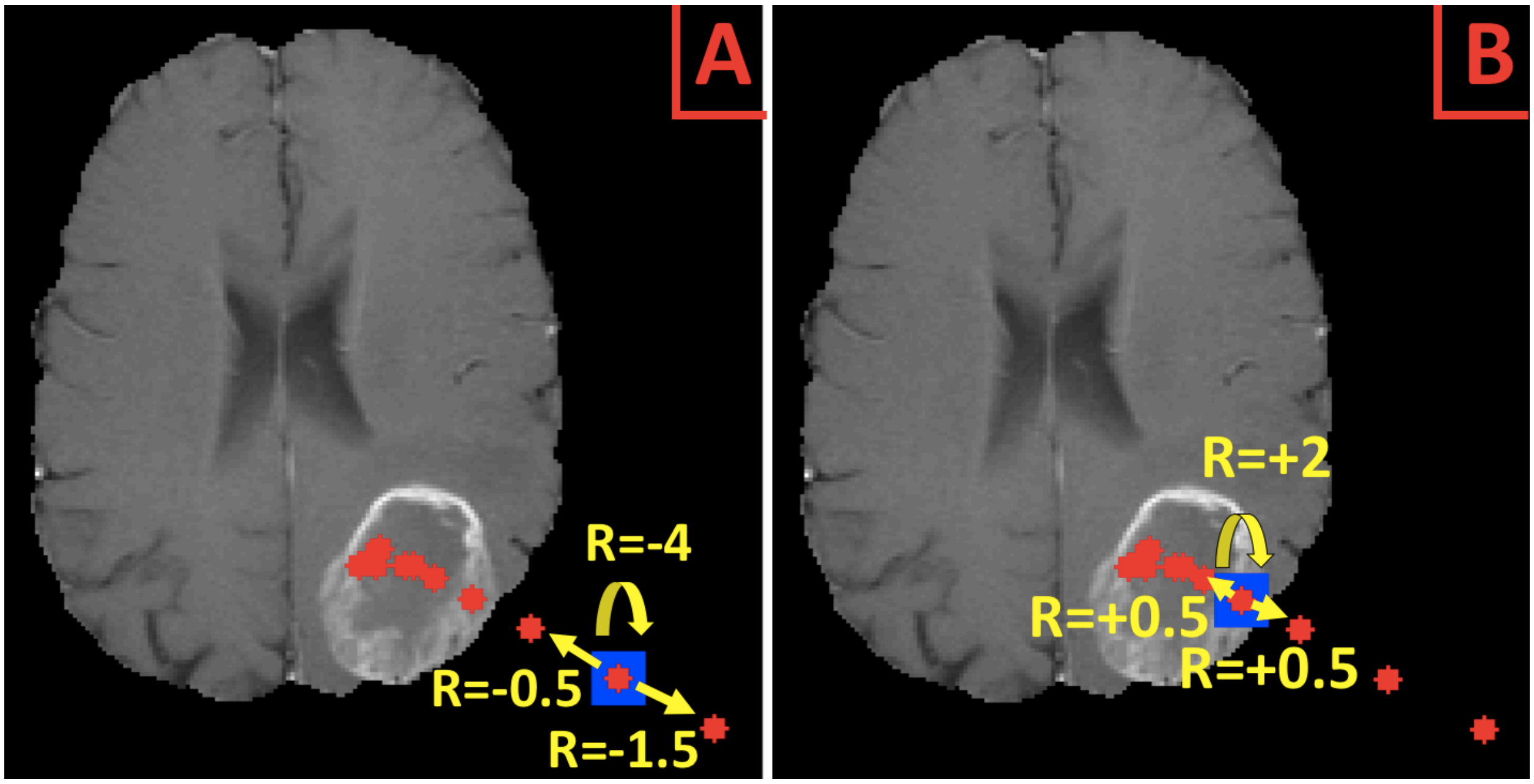

Figure 1: illustration of the reward structure. Figure 1A shows the state in which the agent is located at the second gaze point. The reward of -4 for staying still while not within the lesion penalizes this possible action. Moving forward is rewarded, while moving backward is penalized, but not by as much as staying still. Figure 1B shows a state in which the agent is within the lesion, and is now incentivized by a +2 reward to stay in the same position, so that residing within lesions is favored. However, there are small positive rewards for anterograde and retrograde movement, less than staying still but not a large penalty, so that other states possibly in the lesion can be explored.

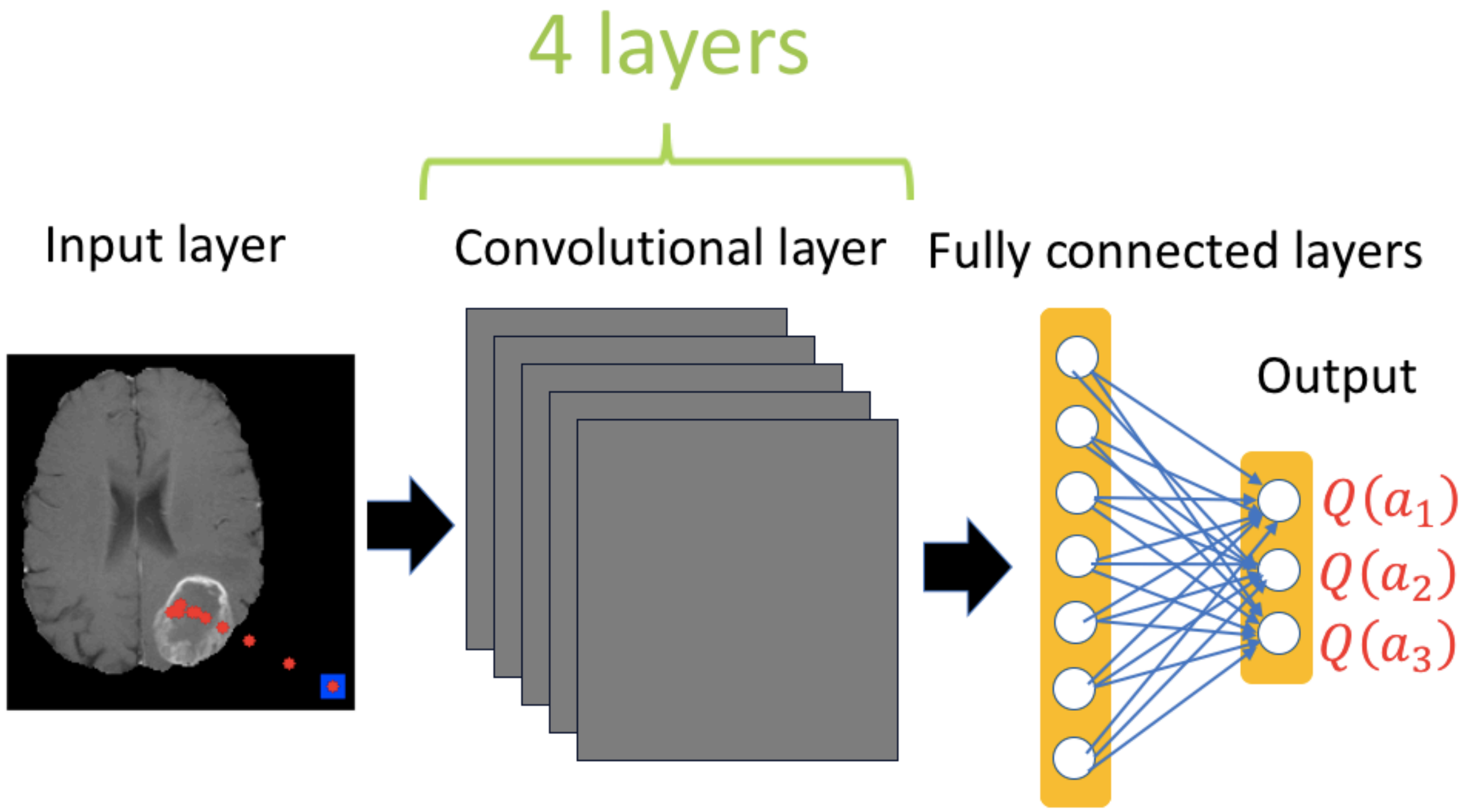

Figure 2: architecture of the deep q neural network.

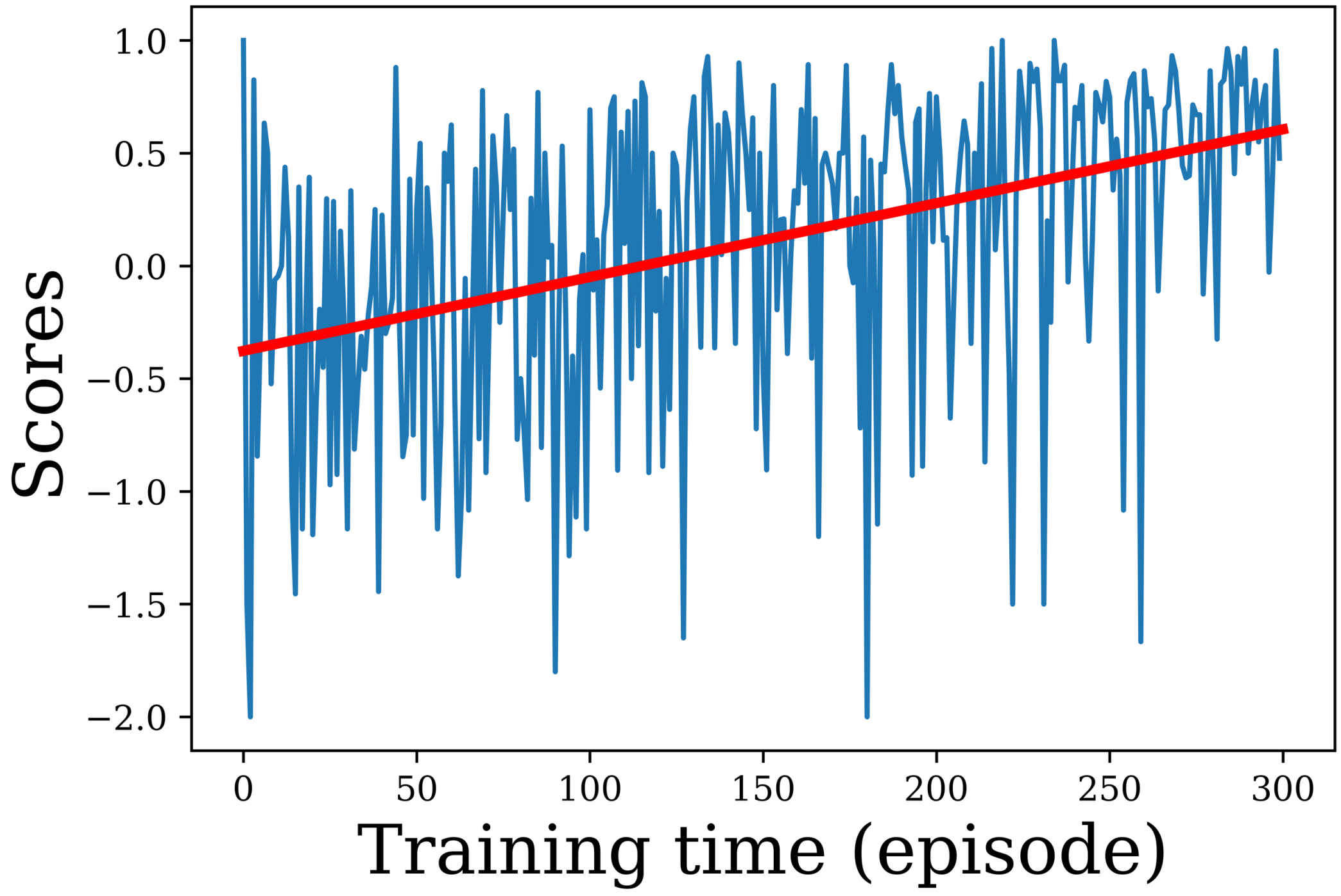

Figure 3: Scores normalized by number of gaze points during the training process of the deep Q network. Although noisy by the nature of the DQN approach, the overall trend is toward increasing accuracy.

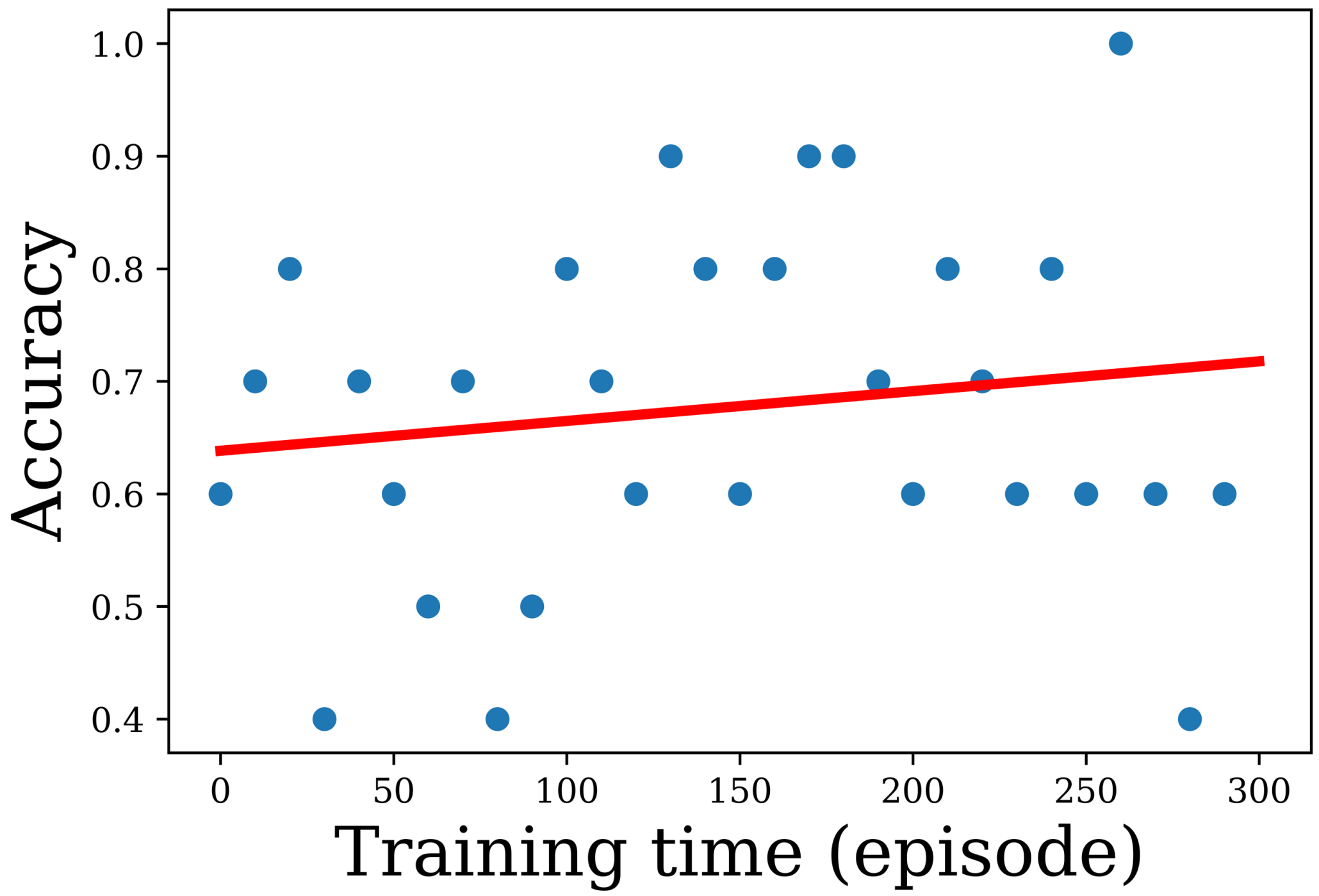

Figure 4: Accuracy of the DQN lesion location prediction on the training set during the course of training, sampled / computed every 10th episode. Although the accuracy is noisy, the overall trend is one of increasing accuracy, as seen by the best fit line with positive slope.

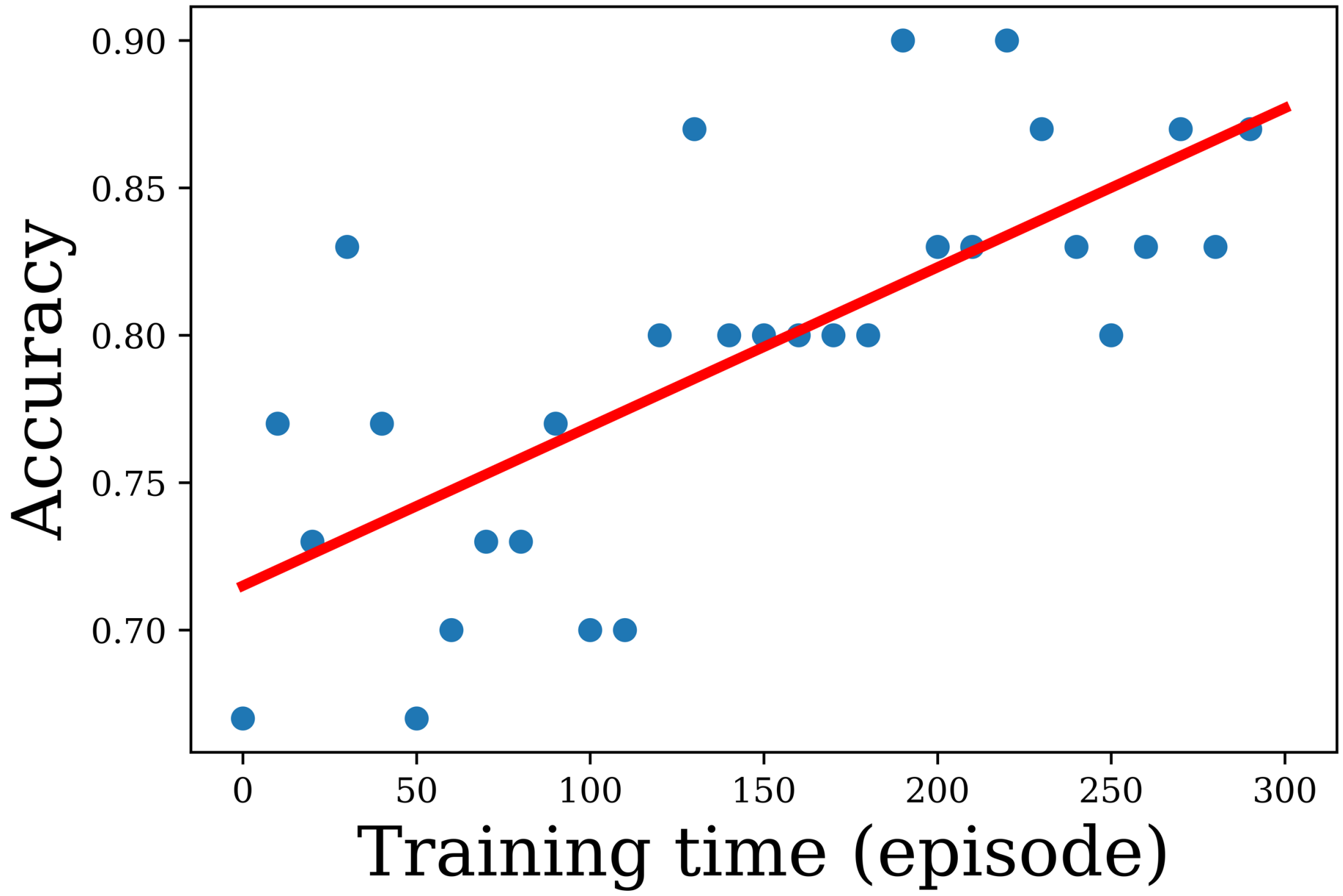

Figure 5: Accuracy of the DQN lesion location prediction on the testing set during the course of training, sampled / computed every 10th episode. Of note is that the results toward the end of training are clustered around 80% accuracy. Additionally, the overall trend is one of improving accuracy, as evidenced by the positive-slope best fit regression line.

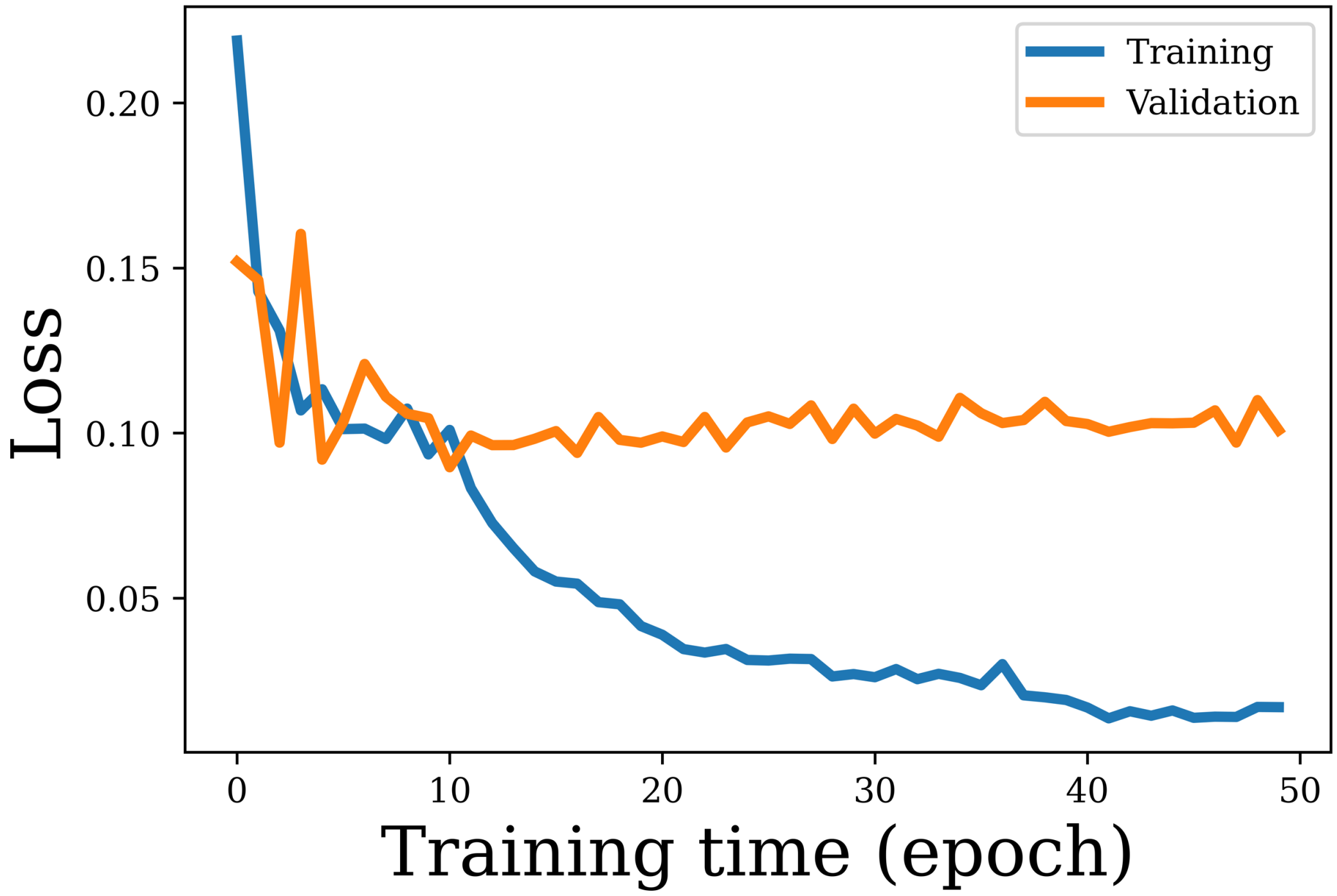

Figure 6: For comparison, we trained a supervised deep learning convolutional neural network (CNN) for 50 epochs using a keypoint detection CNN with overall similar architecture to the network used in our DQN. As the figure shows, before the 10th epoch, the validation set has already diverged and the network is beginning to increasingly overfit the training data.

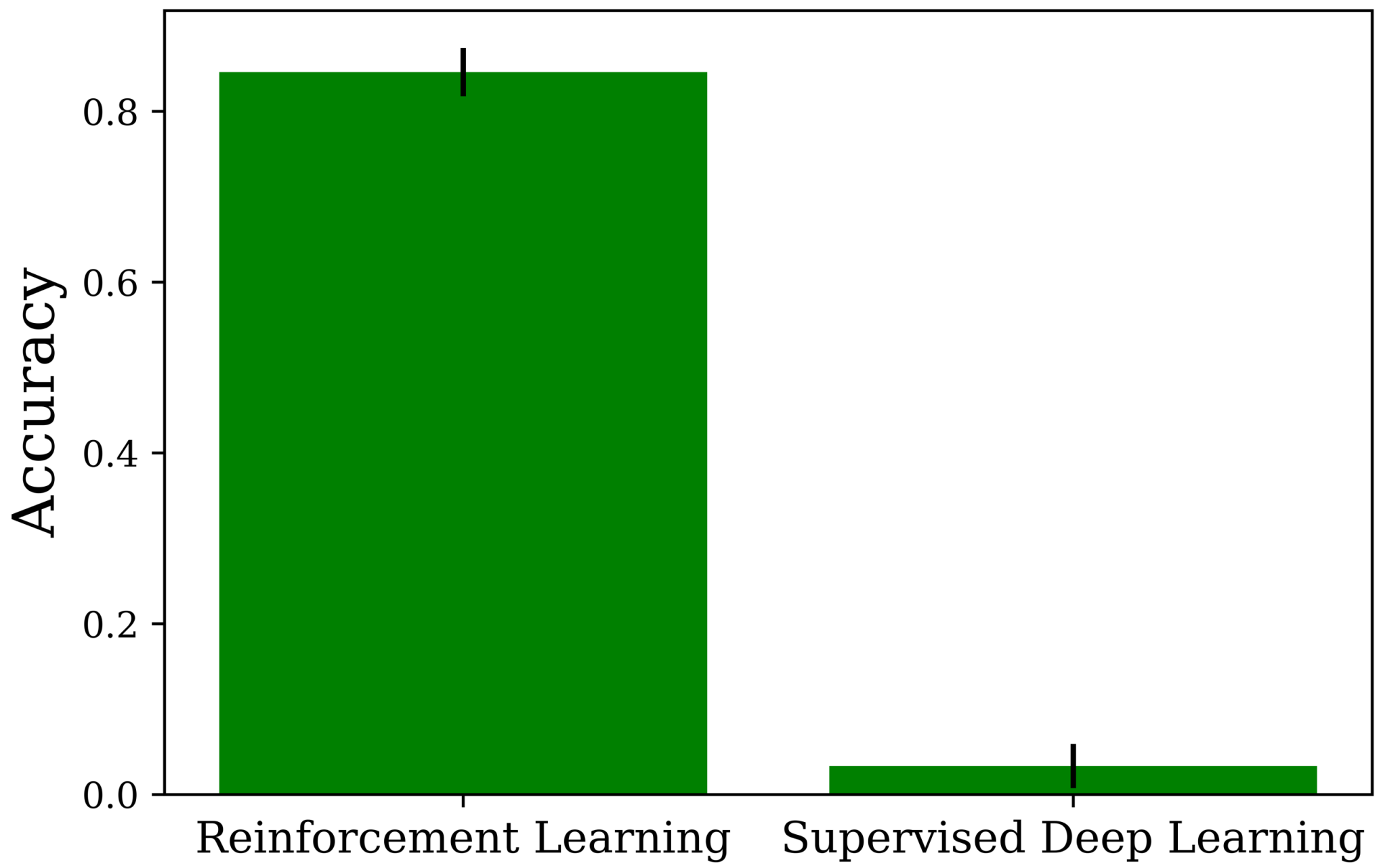

Figure 7: Comparison of the trained reinforcement learning deep Q network predictions on testing set versus those of supervised deep learning. The box heights are average values. Error bars represent standard deviation.